\documentclass[letterpaper, 10 pt, conference]{ieeeconf}  

\IEEEoverridecommandlockouts                              

\usepackage{graphicx}
\usepackage{tabularx}
\usepackage{pgf-pie}
\usepackage{tikz}
\usepackage{xcolor}
\usepackage[dvipsnames]{xcolor}
\usepackage{caption}
\usepackage{subcaption}
\usepackage{float}
\usetikzlibrary{positioning,fit,backgrounds,arrows.meta,calc}
\usepackage{booktabs}
\usepackage{microtype}

\title{\LARGE \bf
Learning to Segment Liquids in Real-world Images
}
\author{Jonas Li, Michelle Li, Luke Liu, Xiaohui Yuan and Heng Fan 
    \thanks{The authors are with the Department of Computer Science and Engineering, University of North Texas, Denton, TX 76203, USA. {\tt\small{chenkuanli@my.unt.edu}}
    }
}

\begin{document}

\maketitle
\thispagestyle{empty}
\pagestyle{empty}

\def\ds{LQDS}
\def\dm{LQDM}

\begin{abstract}

Liquids like water, wine and medicine are everywhere. However, limited attention has been given to the task of segmenting liquids, hindering the ability of robots to safely avoid and interact with them. The segmentation of liquids is difficult because liquids come in diverse appearances and shapes; moreover, they can be both transparent or reflective, taking on arbitrary objects and scenes from their background and surroundings. To take on this challenge, we construct a liquid dataset, LQDS, consisting of 5000 real-world images annotated into 14 distinct classes, and design a novel liquid detection model, LQDM, which leverages cross-attention between a dedicated boundary branch and the main segmentation branch to enhance mask predictions. Extensive experiments demonstrate the effectiveness of LQDM on the testing set of LQDS, outperforming state-of-the-art methods to establish a strong baseline for the semantic segmentation of liquids. We believe that LQDS and LQDM will facilitate future research in liquid segmentation and enable practical applications in robotics. Our dataset and code is released at https://lonaslee.github.io/LQDM/.
\end{abstract}

\section{INTRODUCTION}

Robots operating in real-world environments inevitably encounter liquids, from drinks and cleaning supplies in household settings to puddles and ponds outdoors. The ability to reliably detect and segment liquids is fundamental to safe robot navigation and manipulation: a home-assistant robot must distinguish between beverages to fulfill requests, while a mobile robot must identify spills and wet surfaces to avoid slipping or damaging sensitive payloads. Despite this practical importance, liquid segmentation remains a largely overlooked problem in the robotics vision community.

\begin{figure}[ht!]
  \centering
   \includegraphics[width=1.0\linewidth]{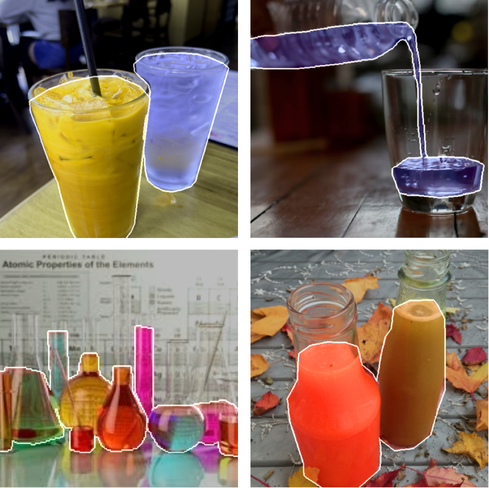}
   \caption{Examples of liquids in real-world settings, highlighting the challenges of liquid segmentation: liquids can reflect their surroundings, exhibit transparency, and deform freely to fit containers of any geometry. Such variability makes distinguishing liquid regions and classes difficult for standard segmentation methods trained on rigid and opaque objects.}
   \label{fig:four-square}
\end{figure}

The challenge of liquid segmentation stems from the unique physical properties of liquids. Unlike the rigid, opaque objects that dominate standard benchmarks, liquids deform freely to fill their containers, producing shapes that range from a coffee mug to a lake. The same liquid type can vary dramatically in color, for instance, orange juice versus cranberry juice. More fundamentally, liquids exhibit unusual optical behavior: some are transparent, revealing the objects behind them, while others are reflective, mirroring their surroundings. These properties cause liquid regions to take on the appearance of whatever is nearby, creating difficulty for models trained on solid object datasets.

Existing liquid datasets fall short of addressing this challenge, covering only synthetic imagery or single liquid types in constrained conditions, leaving a significant gap for the breadth of liquids a robot might encounter in daily operation. To fill this gap, we introduce what is to the best of our knowledge the first dataset for the segmentation of general liquids, a \textbf{L}i\textbf{q}uid \textbf{D}ata\textbf{s}et we call \ds{}, consisting of 5,000 real-world images each with dense, pixel-level masks. LQDS contains 14 distinct classes of liquids, from everyday drinks like coffee and juice to chemicals and miscellaneous liquids. We evaluate state-of-the-art segmentation methods on our testing dataset and find that they perform poorly against liquids, highlighting the need for approaches tailored to the unique properties of liquids.

We propose such a method, a \textbf{L}i\textbf{q}uid \textbf{D}etection \textbf{M}odel which we name LQDM, based on the observation that liquid exhibit higher levels of discontinuity in semantics and low-level features at their boundaries. LQDM’s novel architecture leverages cross-attention between two separate branches to focus on the mask and the boundary, allowing the mask branch to attend to features learned by the boundary branch. By effectively capturing these boundary features, our method outperforms existing methods, establishing the benchmark for the liquid segmentation task. Our contributions are as follows:

\begin{itemize}
    \item We introduce the first general liquid segmentation dataset LQDS, consisting of liquids in diverse scenes from the real world annotated with pixel-level masks.
    \item We propose a novel network \dm{}, which uses cross-attention between parallel branches to inject boundary features into mask predictions, enabling accurate segmentation along challenging liquid edges.
    \item We experimentally validate the effectiveness of \dm{}, attaining higher accuracy on the test set of \ds{} compared to state-of-the-art methods, setting the benchmark and inviting further research on the task.
\end{itemize}

\section{RELATED WORK}

\textbf{Semantic segmentation.} Semantic segmentation assigns a class label to every pixel in an image, and has been a core perception task in robotics for decades. Early deep learning approaches used fully convolutional networks (FCNs)~\cite{FCNs}, treating segmentation as per-pixel classification. Subsequent CNN-based architectures expanded on this foundation: DeepLab~\cite{DL} enlarged the receptive field through dilated convolutions, PSPNet~\cite{PSPNet} introduced pyramid pooling to aggregate multi-scale context, and APCNet~\cite{APCNet} and CCNet~\cite{CCNet} further improved global context modeling, respectively through adaptive pyramids and attention mechanisms. The introduction of vision transformers (ViTs) shifted the paradigm away from CNNs, with methods like Segmenter~\cite{Segmenter} and SegFormer~\cite{SegFormer} demonstrating strong performance through image-level attention. The mask transformer framework, exemplified by MaskFormer~\cite{MaskFormer} and Mask2Former~\cite{Mask2Former}, moved beyond per-pixel classification by introducing learnable mask queries, achieving state-of-the-art results on standard benchmarks. Following this success, EoMT~\cite{EoMT} simplified the mask transformer architecture by removing its pixel decoder, instead demonstrating that plain ViTs can be queried to produce sufficient segmentation masks. However, these methods are designed and evaluated on datasets dominated by rigid, opaque objects that are consistent in texture and shape. This is in contrast to the transparent, reflective, and shape-shifting nature of liquids, and as a result, we find that they transfer poorly to liquid segmentation. Our \dm{} uses a ViT backbone while introducing a parallel cross-attention fused boundary branch, in contrast to the single-branch design of EoMT, in order to handle these unique characteristics of liquids.

Recently, the Segment Anything Model (SAM)~\cite{SAM} was introduced as a pioneering foundation model. SAM and its successor SAM2~\cite{SAM2} uses large-scale pre-training to enable strong generalization for promptable zero-shot segmentation. While not directly designed for semantic segmentation, several works augment SAM with classification heads~\cite{SSA} or leverage its strong image encoder~\cite{SAM2-UNet} for semantic labeling. We evaluate representative SAM-adapted methods on \ds{} and find that they perform comparably to other state-of-the-art semantic segmentation methods, suggesting that SAM's general-purpose features, while powerful, do not inherently address the unique optical challenges presented by liquids. More recently, further zero-shot foundation models including Grounded SAM2~\cite{GroundedSAM2} and SAM3~\cite{SAM3} leverage open-vocabulary grounding to enable text-prompted segmentation. We evaluate SAM3 on \ds{} in a zero-shot setting with class names as text prompts and find that it achieves an IoU of 16\%, suggesting that supervised training on domain-specific data remains essential for liquid segmentation.

\begin{figure}[t!]
\newlength\imw
\setlength{\imw}{0.065\textwidth}
\newcommand{\imgsep}{-0.1em}
\captionsetup[sub]{labelformat=empty, justification=centering, singlelinecheck=false, font=footnotesize}
\newcommand{\lab}[2]{\parbox{1.05cm}{\centering (#1)\\#2}}
\centering
\subcaptionbox{\lab{a}{Water}}{\includegraphics[width=\imw]{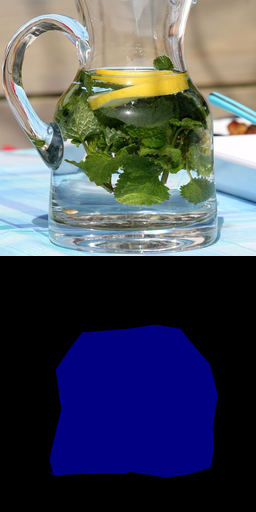}}\hspace{\imgsep}
\subcaptionbox{\lab{b}{Wine}}{\includegraphics[width=\imw]{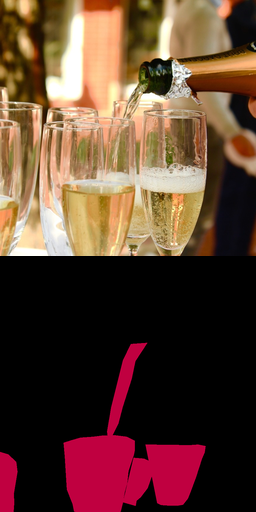}}\hspace{\imgsep}
\subcaptionbox{\lab{c}{Juice}}{\includegraphics[width=\imw]{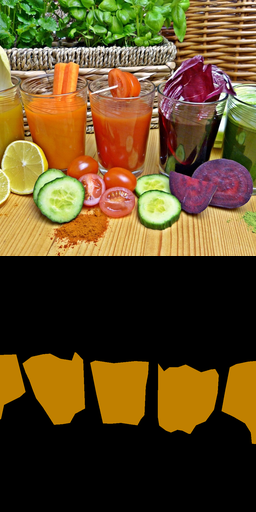}}\hspace{\imgsep}
\subcaptionbox{\lab{d}{Cocktails}}{\includegraphics[width=\imw]{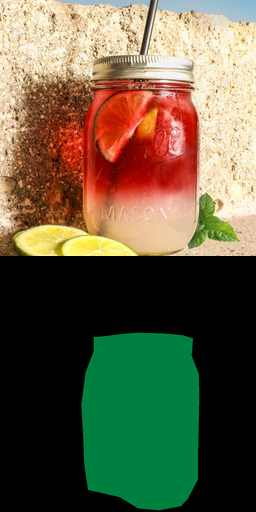}}\hspace{\imgsep}
\subcaptionbox{\lab{e}{Soda}}{\includegraphics[width=\imw]{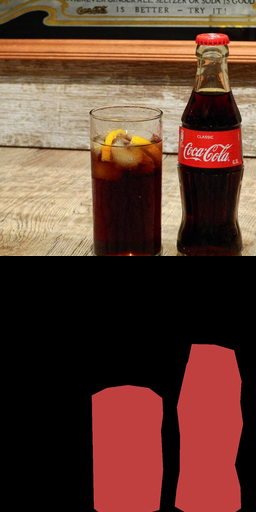}}\hspace{\imgsep}
\subcaptionbox{\lab{f}{Coffee}}{\includegraphics[width=\imw]{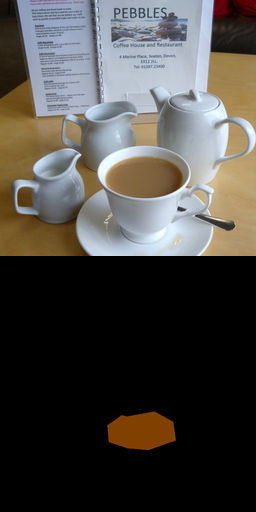}}\hspace{\imgsep}
\subcaptionbox{\lab{g}{Tea}}{\includegraphics[width=\imw]{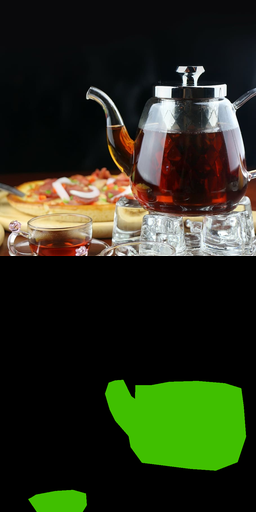}}

\vspace{0.5em}

\subcaptionbox{\lab{h}{Boba}}{\includegraphics[width=\imw]{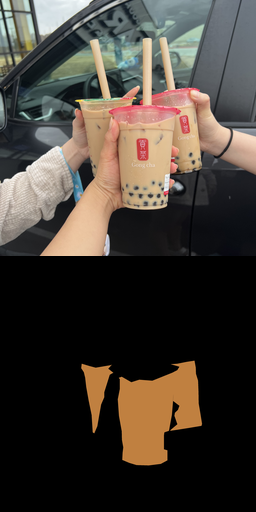}}\hspace{\imgsep}
\subcaptionbox{\lab{i}{Chemical}}{\includegraphics[width=\imw]{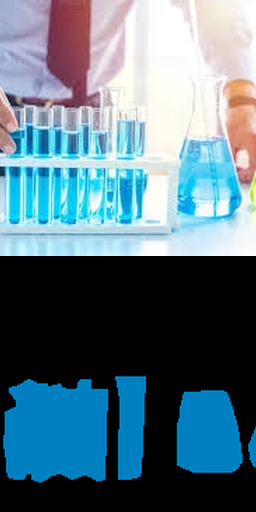}}\hspace{\imgsep}
\subcaptionbox{\lab{j}{Medical}}{\includegraphics[width=\imw]{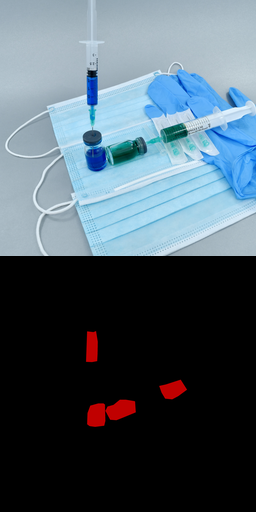}}\hspace{\imgsep}
\subcaptionbox{\lab{k}{Milk}}{\includegraphics[width=\imw]{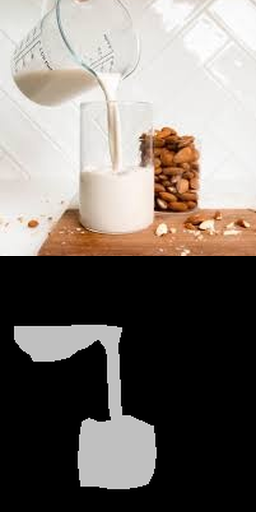}}\hspace{\imgsep}
\subcaptionbox{\lab{l}{Beer}}{\includegraphics[width=\imw]{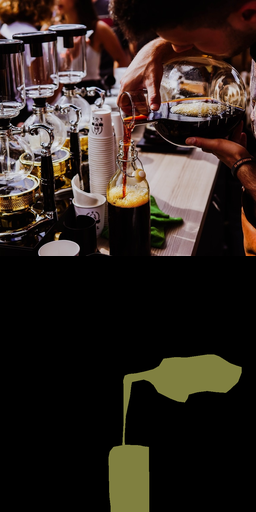}}\hspace{\imgsep}
\subcaptionbox{\lab{m}{Honey}}{\includegraphics[width=\imw]{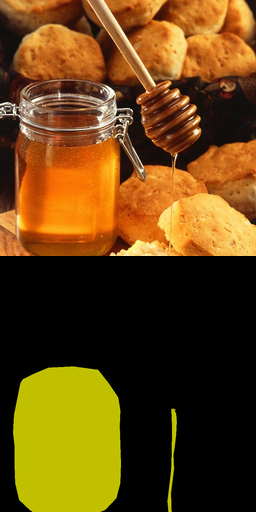}}\hspace{\imgsep}
\subcaptionbox{\lab{n}{Misc}}{\includegraphics[width=\imw]{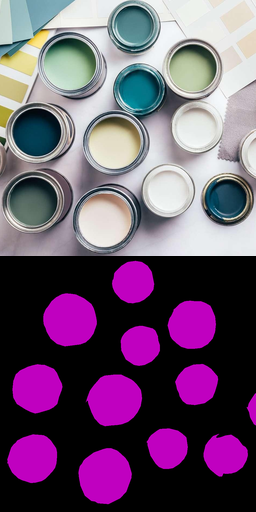}}

\caption{Examples of pairs of images and masks for each of the 14 classes in the \ds{} dataset, demonstrating the variety of forms and appearances that liquids come in.}
\label{fig:classexamples}
\end{figure}

\textbf{Water and chemical segmentation.} Out of all liquids, water has received the most attention. The UW Liquid Pouring Dataset~\cite{UWLiquidPouring}, while large at 4.5 million images, is entirely synthetic, generated from just 81 computer simulations with limited visual diversity. LCDTC~\cite{LCDTC} contains 6K real images, however liquids are limited to residing inside a container, and only bounding boxes of the container are provided. In~\cite{WaterNet}, Liang et al. construct a WaterDataset of natural water bodies like rivers and ponds, applying FCNs to segment them. However, their model relies heavily on dataset-specific priors about water's appearance, limiting its ability to generalize. More recently, Joo et al.~\cite{Habaek} achieve strong water segmentation results by fine-tuning a pretrained SegFormer~\cite{SegFormer}, demonstrating that transformer-based models can implicitly learn appearance priors given sufficient data. For chemicals, Schober et al.~\cite{LabLiquidVision} construct the LabLiquidVolume dataset, while Eppel et al.~\cite{LabPics2} construct the LabPics2 dataset, consisting of 5K and 8K images, respectively. However, LabLiquidVolume is constrained to only 12 predefined containers and 3 predefined colors, and LabPics2 provides only a single class for liquids. Both works apply CNN methods for segmentation, and although effective in their respective domains, these works frame liquid detection as a binary classification problem for a single liquid type, relying on narrow appearance assumptions that do not generalize across the diversity of liquids a robot encounters in practice.

\textbf{Transparent and reflective object segmentation.} Liquids share optical properties with other challenging object categories that have been more widely studied. For mirror segmentation, MirrorNet~\cite{MirrorNet} exploits semantic and low-level discontinuities at mirror boundaries, extracted at multiple scales using dilated convolutions. For glass segmentation, GDNet~\cite{GDNet} leverages large-scale contextual cues through separable convolutions with varying kernel and dilation sizes. TransLab~\cite{TransLab} and Trans2Seg~\cite{Trans2Seg} tackle general transparent object segmentation, the latter treating it as a multi-class problem distinguishing between object categories, a closer formulation to our own.

However, these methods are not well-suited to liquids, as not all liquids are transparent or reflective, and those which are often behave differently from glass or mirrors. For instance, honey is semi-transparent, but introduces color distortion and refraction that glass-based methods do not anticipate. Wine, water, and certain chemicals exhibit surface reflectance governed by the Fresnel effect, which differs fundamentally from planar mirror reflection. The same liquid can even shift between transparency, reflectance, and opacity depending on viewing angle and lighting. A robot operating across varied environments needs a model that handles this full range of behavior, rather than one optimized for a single optical property.

To our knowledge, no prior work addresses the semantic segmentation of general liquids. We aim to close this gap using both a comprehensive dataset and a model designed to handle the unique and varied challenges presented by liquids.

\section{Liquid Segmentation Dataset}

We construct \ds{}, the first dataset for general liquid segmentation, to address the lack of suitable training and evaluation data for this task. \ds{} contains 5,000 real-world images with dense pixel-level annotations across 14 liquid classes.

\textbf{Dataset Description.} Images in \ds{} are gathered using both the internet and original photographs captured with phone cameras in everyday environments. This combination yields wide variation in scale, viewpoint, lighting, and background, representative of what a robot would encounter in real deployment. The 14 liquid classes are \textit{water, wine, juice, cocktails, soda, coffee, tea, boba, chemical, medical, milk, beer, honey, and miscellaneous}, reflecting a wide range of liquids robots need to identify in household, commercial, and outdoor settings. For training and evaluation, we perform a stratified split with random sampling that preserves class proportions across both sets, yielding 4,200 training images and 800 test images.

\textbf{Annotation.} For high-quality annotation, each image in our dataset is manually labeled by experts (\textit{e.g.}, students who work on related fields). We annotate only liquid itself; container walls, straws, spoons, and other objects present within or around the liquid are excluded, as seen in Figure~\ref{fig:classexamples}. A single annotator assigns all class labels at the end, which ensures consistency across categories but precludes measurement of inter-annotator agreement. To further ensure annotation quality, we perform inspection and refinement until the annotation of every image meets the requirement. 

\begin{figure}[t!]
    \centering
    \begin{subfigure}[t]{0.48\columnwidth}
        \centering
        \includegraphics[width=\linewidth]{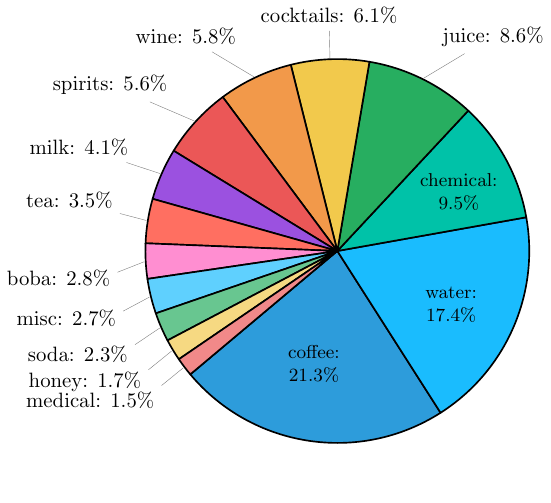}
        \caption{Object count distribution by class}
        \label{subfig:dataset-pie}
    \end{subfigure}%
    \hfill
    \begin{subfigure}[t]{0.48\columnwidth}
        \centering
        \begin{tikzpicture}
            \node(anchor)[]{};
            \node(train)[left=-.2cm of anchor]{
                \includegraphics[width=1.8cm]{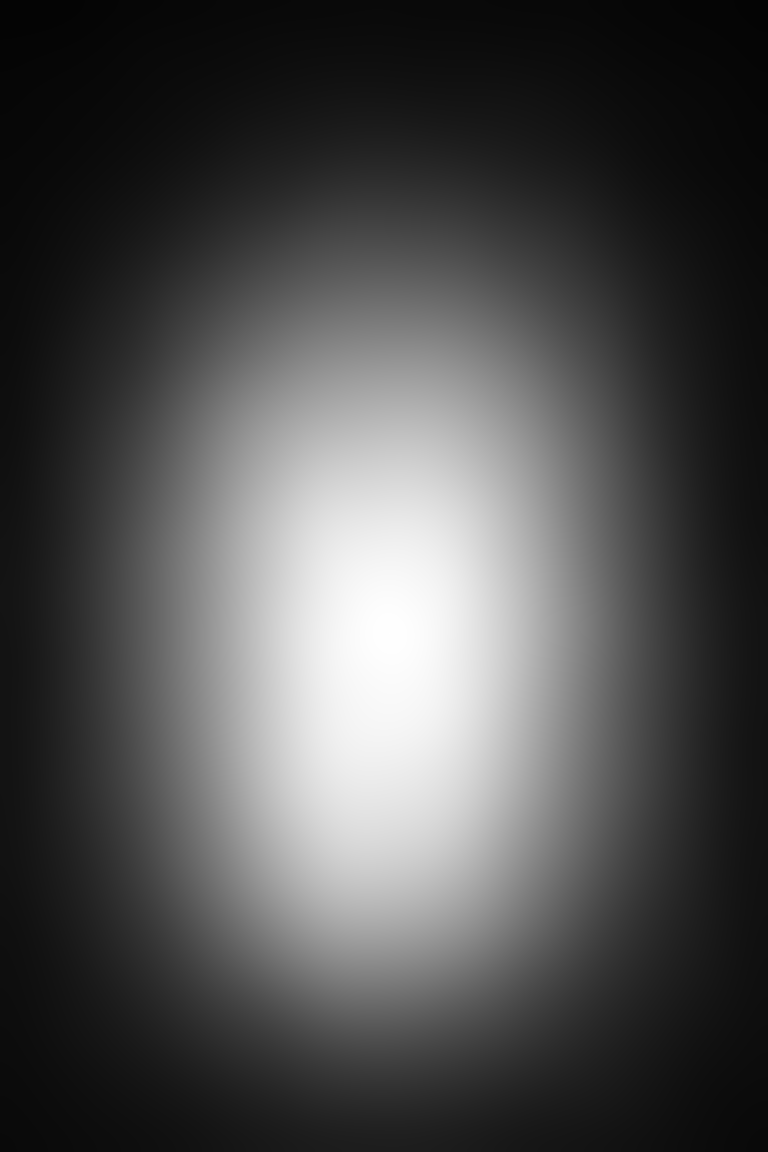}
            };
            \node[above=-.2cm of train]{\small{Training Set}};
            \node(test)[right=-.2cm of anchor]{
                \includegraphics[width=1.8cm]{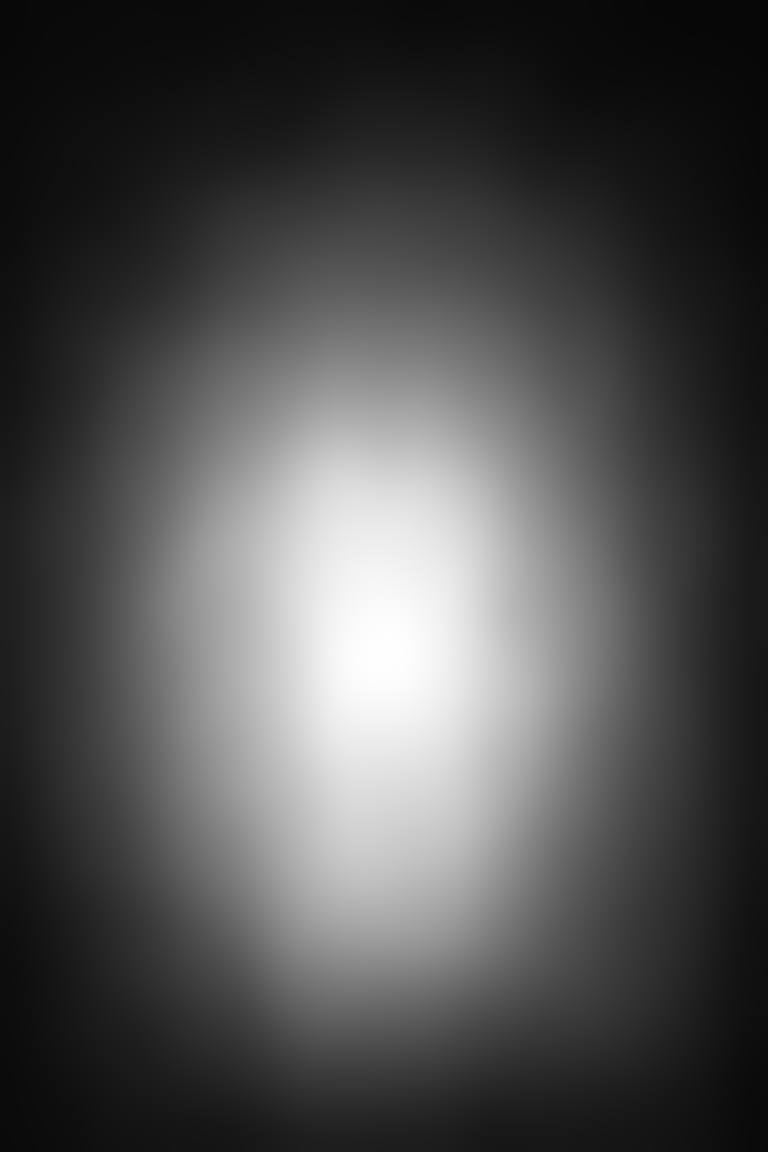}
            };
            \node[above=-.2cm of test]{\small{Testing Set}};
            \node(belowexpander)[below=0 of train]{};
        \end{tikzpicture}
        \caption{Liquid location distribution.}
        \label{subfig:dataset-location}
    \end{subfigure}
    
    \vspace{0.5em}

    \begin{subfigure}[t]{0.48\columnwidth}
        \centering
        \includegraphics[width=\linewidth]{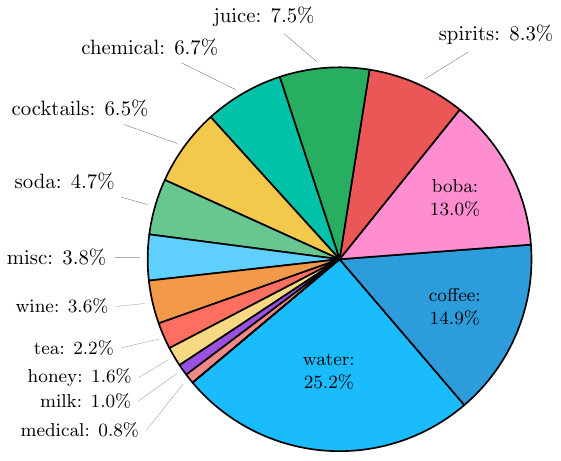}
        \caption{Pixel distribution by class}
        \label{subfig:dataset-pixel-pie}
    \end{subfigure}
    \hfill
    \begin{subfigure}[t]{0.48\columnwidth}
        \centering
        \includegraphics[width=\linewidth]{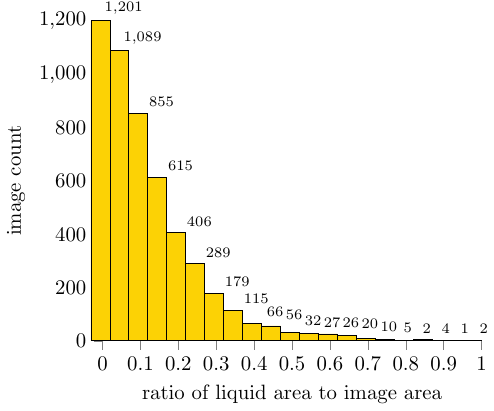}
        \caption{Liquid area distribution}
        \label{subfig:dataset-area}
    \end{subfigure}%
    \caption{Statistics of the \ds{} dataset.}
    \label{fig:dataset-stats}
\end{figure}

\textbf{Dataset Analysis.} Figure~\ref{fig:dataset-stats} summarizes the key properties of \ds{}.

\begin{itemize}
    \item \textbf{Object count distribution.} Figure~\ref{subfig:dataset-pie} shows the percentage of images in which each class appears at least once. There are 117 training images and 21 test images which contain instances of more than one liquid class simultaneously. Coffee appears the most, followed by water, chemicals, juice, and then the alcoholic beverages. 
    
    \item \textbf{Pixel distribution.} Figure~\ref{subfig:dataset-pixel-pie} shows each class's share of total annotated pixels. Despite appearing less frequently than coffee, water accounts for the largest pixel share, consistent with water's tendency to appear in large, uncontained bodies such as ponds and pools rather than small cups or glasses.
    
    \item \textbf{Liquid location.} Figure~\ref{subfig:dataset-location} shows spatial probability maps of liquid pixels within images. Liquids tend toward the image center in both the training and test sets, reflecting the tendency for photographers and robot cameras to center the subject of interest.
    
    \item \textbf{Liquid area.} Figure~\ref{subfig:dataset-area} shows the distribution of the percentage of each image occupied by liquid. LQDS contains liquids of varying sizes, with most falling below the 20\% mark. These correspond to drinks in cups, bottles, and glasses, which are the liquids most commonly encountered by household and service robots. Above the 20\% mark consists of both larger liquid bodies, such as fish tanks and pools, as well as close-up shots where liquid fills most of the frame.
\end{itemize}

\begin{figure*}[ht!]
    \centering
    \includegraphics[width=1\linewidth]{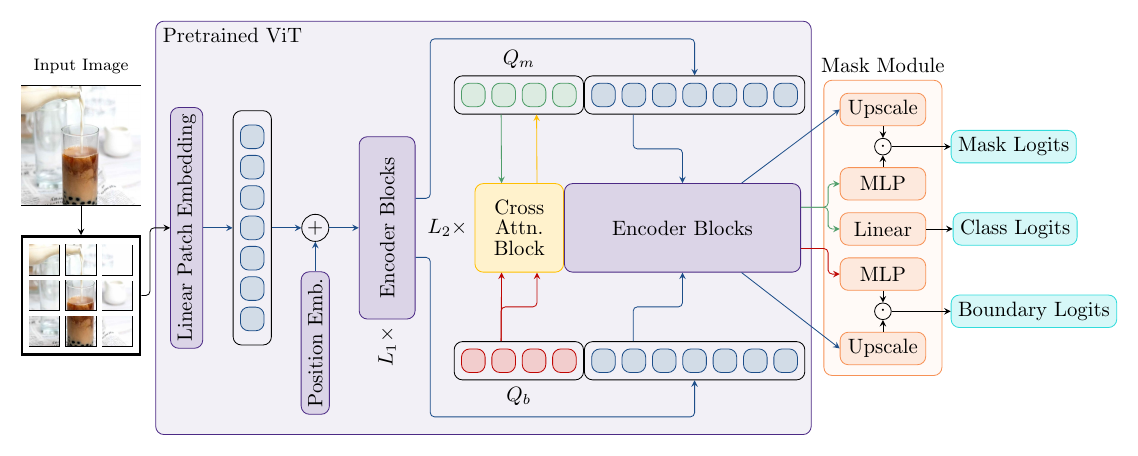}
    \caption{The overall architecture of \dm{}. The network splits into two branches: the \textit{mask branch} for semantic segmentation and the \textit{boundary branch} for boundary-enhanced feature integration.}
    \label{fig:arch}
\end{figure*}

\section{Proposed Method}
We observe that there tends to be higher levels of discontinuity in content and low-level color or texture at the boundary of liquids, and we argue that these boundaries are easier to detect due to this contrast. We take advantage of this by proposing a dual-branch architecture to inject boundary features into mask predictions through cross-attention.

\subsection{Overview}

Figure~\ref{fig:arch} illustrates our architecture, \dm{}, consisting of two distinct output layers that facilitate multi-task learning. For a single image, feature embeddings are extracted by a vision transformer (ViT) backbone using $L_1$ transformer encoder blocks. Following these shared blocks, we split into two separate branches: the mask branch and the boundary branch. For each branch, we introduce a set of $K$ learnable queries. These queries are processed by $L_2$ cross-attention blocks, allowing the mask branch to attend to the boundary branch. Following cross-attention, the queries are concatenated onto the feature embeddings of their respective branch, and together they are processed by $L_2$ transformer encoder blocks. Notably, these transformer encoder blocks are shared between the two branches, allowing the ViT backbone itself to learn the boundary features. We discover this leads to better performance while being more parameter efficient and taking up less memory during training.

To produce the final logits for prediction, a mask module is installed to process the query and feature embedding outputs of the last transformer encoder block. The class logits are acquired using a linear layer, while the mask and boundary logits are acquired using multilayer perceptrons followed by dot products with upscaled image features.

\subsection{Boundary Attention}
\label{subsec:cab}

To leverage boundary information for improved mask prediction, cross-attention blocks are introduced alongside each transformer encoder block, allowing the mask branch to attend to features from the boundary branch. Specifically, at every $i$-th pair of cross-attention and transformer encoder blocks, we take the two sets of $K$ learnable queries from both the mask branch and the boundary branch and apply cross-attention between them, with the mask queries $Q_m^{(i)}$ serving as the query and the boundary queries at the same block $Q_b^{(i)}$ providing the key and value pairs:
\begin{equation}
    Q_m^{(i)} = Q_m^{(i)} + CrossAttnBlock(Q_m^{(i)}, Q_b^{(i)})
    \label{eq:cab}
\end{equation}
where $CrossAttnBlock$ implements standard multi-head cross-attention. We normalize both $Q_m$ and $Q_b$ using layer normalization before producing the queries and key-value pairs from them using projection layers. The cross-attention block is applied alongside each of the $L_2$ transformer blocks, allowing progressive refinement of mask predictions guided by boundary features. Our cross-attention mechanism allows the network to selectively attend to relevant boundary features at different spatial locations, leading to more accurate segmentation especially along liquid edges where transparency and reflection create ambiguities.

\subsection{Mask Module}

We use a mask module as per \cite{Mask2Former, EoMT} to obtain class, mask, and boundary logits. The mask module takes as inputs the corresponding queries and image feature embeddings of the mask and boundary branch following the last of the $L_2$ cross-attention and transformer encoder blocks. For the class logits, the queries are passed through a simple linear layer. For the mask and boundary logits, the respective queries are passed first through a three-layer MLP, and the logits are then produced from a dot product with the upscaled image feature embeddings.

\begin{figure*}
  \centering
   \includegraphics[width=1.0\linewidth]{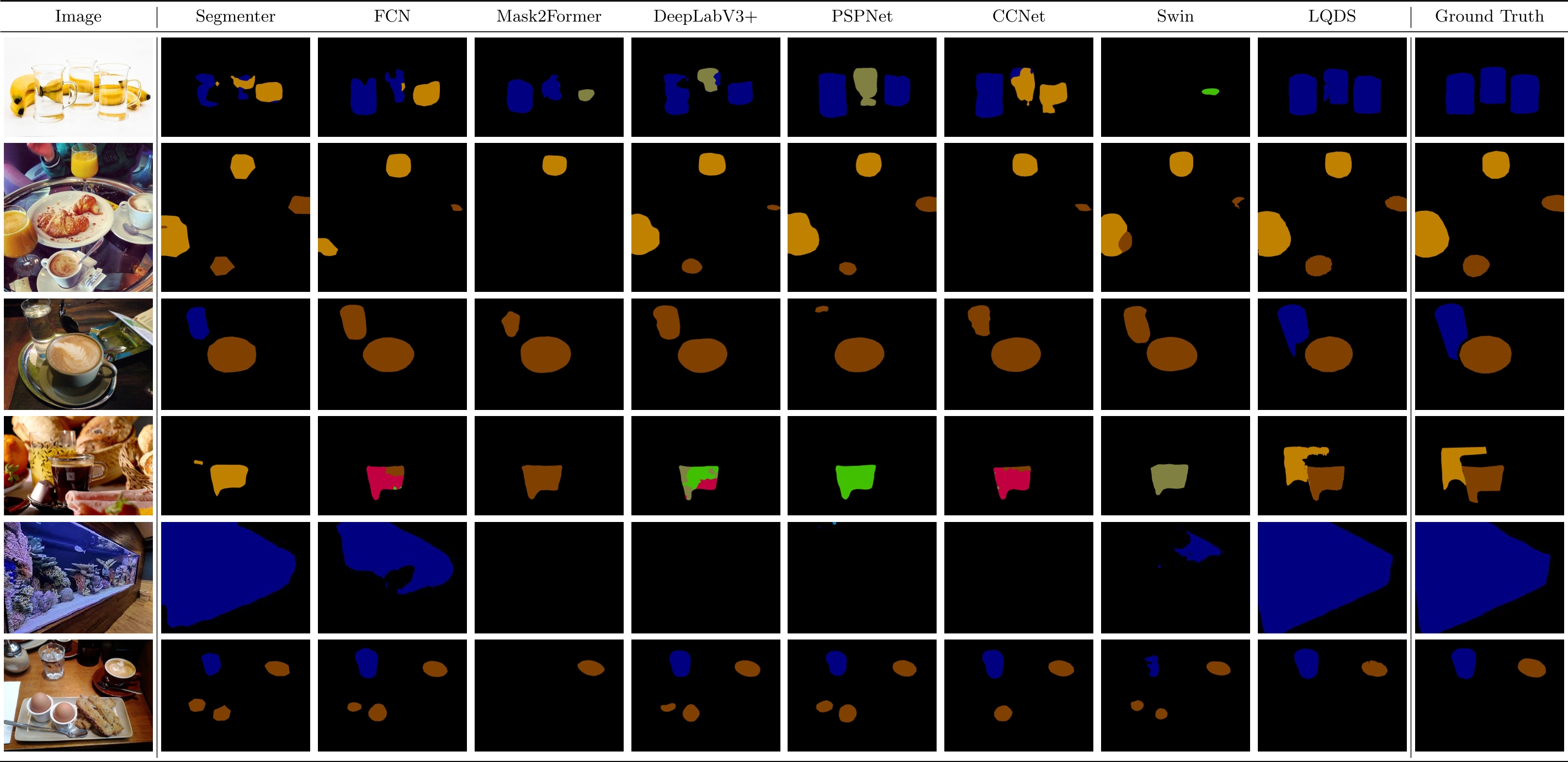}
   \caption{Visual comparison of \dm{} to other semantic segmentation methods on images from the \ds{} testing set.}
   \label{fig:mask-compare}
\end{figure*}

\begin{figure}[!t]
    \centering
    \begin{subfigure}[t]{0.48\columnwidth}
        \centering
        \includegraphics[width=\linewidth]{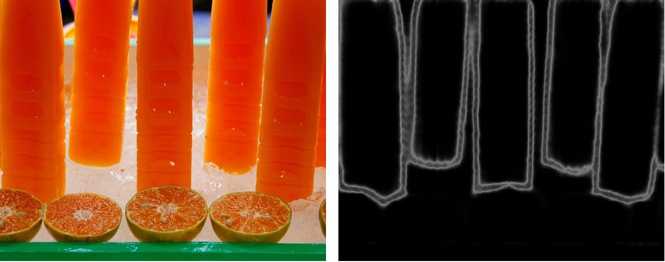}
    \end{subfigure}%
    \hspace{.1em}
    \begin{subfigure}[t]{0.48\columnwidth}
        \centering
        \includegraphics[width=\linewidth]{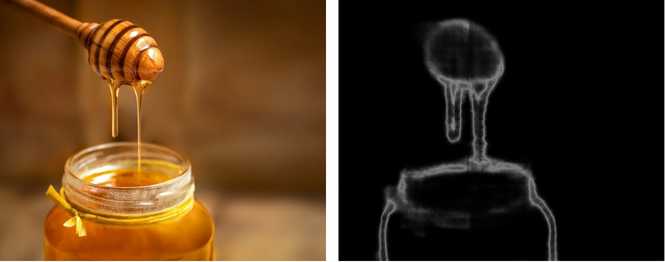}
    \end{subfigure}%
    
    \vspace{0.1em}
    
    \begin{subfigure}[t]{0.48\columnwidth}
        \centering
        \includegraphics[width=\linewidth]{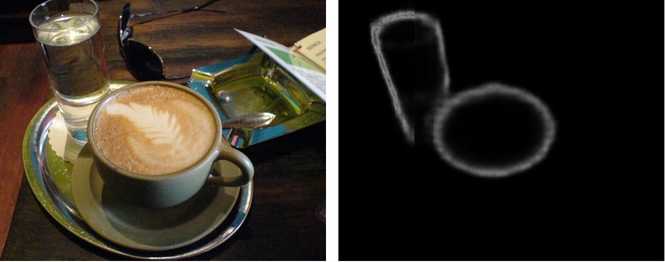}
    \end{subfigure}%
    \hspace{.1em}
    \begin{subfigure}[t]{0.48\columnwidth}
        \centering
        \includegraphics[width=\linewidth]{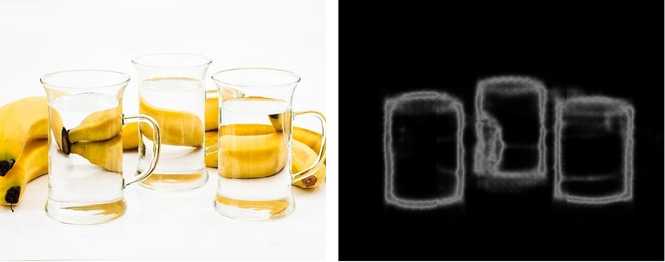}
    \end{subfigure}

    \vspace{0.1em}
    
    \begin{subfigure}[t]{0.48\columnwidth}
        \centering
        \includegraphics[width=\linewidth]{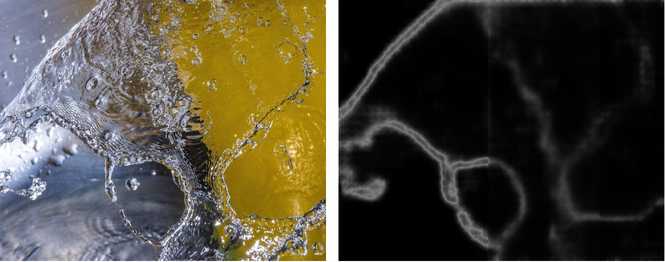}
    \end{subfigure}%
    \hspace{.1em}
    \begin{subfigure}[t]{0.48\columnwidth}
        \centering
        \includegraphics[width=\linewidth]{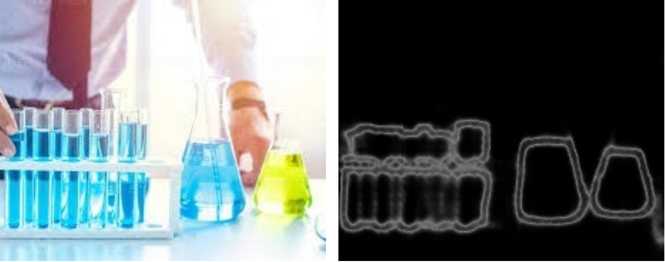}
    \end{subfigure}
    \caption{Visualization of predicted boundaries by \dm{}.}
    \label{fig:boundvis}
\end{figure}

\subsection{Loss Function}

The segmentation and boundary logits obtained from the mask module are used to compute overall training loss. In order to supervise both branches during training, we define the total loss as the weighted sum of the mask branch loss and the boundary branch loss:
\begin{equation}
    L = L_m + \omega L_b
    \label{eq:loss}
\end{equation}
where $L_m$ and $L_b$ denote the mask and boundary losses, balanced by the parameter $\omega$. Dice Cross-Entropy and Binary Cross-Entropy (BCE) are used to calculate $L_m$ and $L_b$, respectively. Boundary BCE s computed against boundary ground-truths generated as a binary mask with a thickness of 1\% of the image diagonal length (computed as $\sqrt{H^2+W^2}$).

\section{Experiments}

\subsection{Implementation Details}

We implement \dm{} using PyTorch. For training, we use a $512 \times 512$ input size into the model, augmented using a sequence of color jittering, random horizontal flipping, scale jittering, padding, and random cropping. For the feature extraction backbone, we choose a pretrained SigLIP2 backbone~\cite{SigLIP2}, and use $L_1=20$, $L_2=8$. The other parts of our network are initialized randomly. For the multi-head cross-attention operation, we use 8 attention heads with a dropout rate of 0.2. For loss optimization, we use the AdamW optimizer with a learning rate of $10^{-4}$ and layer-wise learning rate decay of 0.8 for 32 epochs. We train on four NVIDIA A5500 GPUs with a batch size of four per GPU. On this hardware, it takes about six hours for the network to converge. For inference, the shorter side of the image is resized to the input size $512$, and the model processes the image in squares using a sliding-window fashion along the longer side.

\textbf{Evaluation metrics.} For a comprehensive evaluation, we adopt the Intersection over Union (IoU) and pixel accuracy (PA) metrics standard for image segmentation. For both the mean IoU and mean PA, we only factor in the values from the object classes, ignoring background.

\begin{table*}[]
\centering
\caption{Per-class IoU, mean IoU, and mean PA performance on the testing set of \ds{} for evaluated segmentation methods, sorted by mean IoU.}
\label{tab:classIoU}
\resizebox{\textwidth}{!}{\begin{tabular}{l|cccccccccccccc|cc}
    \toprule
    Model & water & wine & juice & cocktails & soda & coffee & tea & boba & chemical & medical & milk & beer & honey & misc & mIoU $\uparrow$ & mPA \\
    \midrule
    KNet & 21.01 & 10.87 & 28.94 & 8.45 & 13.09 & 39.97 & 2.93 & 56.42 & 15.80 & 0.74 & 1.08 & 20.04 & 17.54 & 23.27 & 18.58 & 45.95 \\
    TransLab & 35.94 & 11.63 & 42.73 & 2.84 & 15.50 & 40.52 & 0.12 & 49.13 & 25.78 & 0.00 & 38.63 & 28.08 & 21.52 & 4.20 & 22.62 & 38.25 \\
    GCNet & 28.80 & 13.11 & 33.88 & 39.78 & 11.63 & 47.11 & 10.26 & 46.49 & 16.53 & 0.00 & 23.53 & 33.35 & 15.73 & 32.37 & 25.18 & 45.79 \\
    SAM2-Unet & 66.90 & 26.50 & 42.30 & 36.50 & 58.20 & 59.90 & 8.30 & 67.30 & 41.30 & 7.40 & 34.30 & 27.60 & 39.50 & 22.10 & 38.44 & 57.40 \\
    SSA & 55.21 & 26.76 & 51.77 & 30.69 & 48.89 & 59.41 & 15.28 & 73.40 & 27.37 & 35.29 & 59.88 & 50.37 & 22.73 & 19.49 & 41.18 & 58.21 \\
    ANN & 59.40 & 14.92 & 44.85 & 33.79 & 53.83 & 59.10 & 14.37 & 63.86 & 43.63 & 19.13 & 39.26 & 46.13 & 63.97 & 33.44 & 42.12 & 53.27 \\
    Segmenter & 54.14 & 16.36 & 48.22 & 44.92 & 42.91 & 63.55 & 15.94 & 62.93 & 37.90 & 24.66 & 50.94 & 50.50 & 51.51 & 37.25 & 42.98 & 53.44 \\
    FCN & 44.10 & 16.46 & 48.63 & 45.42 & 52.38 & 61.45 & 13.65 & 62.88 & 42.07 & 23.30 & 52.31 & 40.27 & 62.68 & 37.11 & 43.05 & 53.48 \\
    PSPNet & 50.54 & 21.12 & 48.47 & 43.37 & 56.45 & 59.89 & 13.88 & 61.38 & 44.69 & 13.21 & 50.87 & 44.91 & 60.42 & 42.51 & 43.69 & 55.17 \\
    APCNet & 54.06 & 20.66 & 50.98 & 40.94 & 52.01 & 63.70 & 16.57 & 65.11 & 39.38 & 13.48 & 62.98 & 32.91 & 66.84 & 34.83 & 43.89 & 54.91 \\
    DeepLabV3+ & 53.39 & 16.25 & 50.77 & 34.62 & 48.93 & 61.22 & 12.91 & 68.33 & 45.70 & 24.43 & 48.78 & 46.90 & 65.78 & 39.16 & 44.08 & 55.11 \\
    Mask2former & 49.36 & 15.18 & 50.02 & 39.24 & 45.77 & 60.67 & 8.93 & 65.81 & 30.35 & 22.03 & 60.86 & 51.64 & 70.81 & \textbf{46.66} & 44.10 & 54.35 \\
    CCNet & 48.67 & 15.35 & 49.74 & \textbf{47.06} & 56.85 & 62.53 & \textbf{21.71} & 66.07 & 45.92 & 25.56 & 47.42 & 46.04 & 59.94 & 35.21 & 44.86 & 56.12 \\
    Maskformer & 43.24 & 18.11 & 50.16 & 45.10 & 60.61 & 62.22 & 16.99 & 70.87 & 42.47 & 18.38 & 59.66 & 48.75 & 56.58 & 37.52 & 45.05 & 55.94 \\
    Segformer & 54.94 & 19.09 & 48.34 & 42.11 & 62.95 & 65.19 & 9.83 & 74.24 & 35.22 & 24.88 & 46.29 & 52.77 & 64.56 & 36.20 & 45.47 & 57.02 \\
    Swin & 40.76 & 15.21 & 51.54 & 43.47 & 56.34 & 66.13 & 12.78 & 76.11 & 39.67 & 31.37 & 67.19 & 46.67 & 60.15 & 34.73 & 45.87 & 56.30 \\
    EoMT & 72.94 & 21.59 & 46.07 & 41.18 & 76.63 & 70.93 & 17.92 & 80.64 & 54.04 & 24.49 & 55.82 & 61.49 & 54.83 & 49.73 & 52.02 & 64.14 \\
    \midrule
    LQDM & \textbf{75.68} & \textbf{35.97} & \textbf{56.41} & 44.00 & \textbf{79.63} & \textbf{74.55} & 16.09 & \textbf{78.63} & \textbf{55.04} & \textbf{59.36} & \textbf{68.50} & \textbf{63.23} & \textbf{78.93} & 43.87 & \textbf{59.28} & \textbf{71.61} \\
    \bottomrule
\end{tabular}}
\end{table*}

\begin{table}[!t]
\caption{Component analysis. BL is the baseline, SL is the SigLIP2 weights, BB is the boundary branch, BCA is the boundary cross attention between the two branches.}
\label{tab:ablation-arch}
\centering
\begin{tabular}{l|cccc}
\toprule
Components & mIoU & mAcc & mPrec & mF1 \\
\midrule
BL & 52.02 & 64.14 & 73.15 & 66.09 \\
BL+SL & 56.50 & 70.28 & 72.41 & 69.03 \\
BL+SL+BB & 57.91 & 71.30 & 74.69 & 70.57 \\
BL+SL+BB+BCA & \textbf{59.28} & \textbf{71.61} & \textbf{76.63} & \textbf{71.50} \\
\bottomrule
\end{tabular}
\end{table}

\subsection{Comparison with State-of-the-Arts}

For our evaluation, we select 17 state-of-the-art methods in semantic segmentation, including both well-established CNN-based models such as DeepLabV3+~\cite{DLV3+}, PSPNet~\cite{PSPNet}, APCNet~\cite{APCNet}, FCN~\cite{FCNs}, ANN~\cite{ANN}, TransLab~\cite{TransLab}, KNet~\cite{KNet}, and CCNet~\cite{CCNet} and transformer-based models such as Swin~\cite{Swin}, MaskFormer~\cite{MaskFormer}, Mask2Former~\cite{Mask2Former}, Segmenter~\cite{Segmenter}, Segformer~\cite{SegFormer}, GCNet~\cite{CCNet}, and EoMT~\cite{EoMT}, as well as recent SAM-based models such as Semantic Segment Anything (SSA)~\cite{SSA} and SAM2-UNet~\cite{SAM2-UNet}. We retrain each of these networks on the training set of \ds{} using their publicly available implementations and parameter settings~\cite{MMSeg}, and evaluate them on the \ds{} testing set.

Figure~\ref{fig:mask-compare} visualizes segmentation masks produced by \dm{} and existing methods. In rows 1 and 2, it can be seen that \dm{} is effective on images where the foreground and background are semantically complex, with many objects intersecting the target region. \dm{} is able to correctly classify the liquids where others mistake it for another class, such as in rows 3 and 4. Additionally, \dm{} identifies the liquids where others fail to see them, such as in row 5, but is also less prone to false positives, as seen in row 6.

Table~\ref{tab:classIoU} reports the per-class liquid segmentation performance on the testing set of \ds{}, and it can be seen that \dm{} achieves the highest mean IoU score of 59.28\% and the highest mean PA of 71.61\%.

\subsection{Ablation Study}

\textbf{Architecture components}. Table~\ref{tab:ablation-arch} evaluates the effectiveness of our proposed method. For this part, in addition to the IoU and PA metrics, we calculate the pixel precision and F1 score. For our baseline, we use EoMT with a DINOv2 backbone. Through experimentation we find that initializing the encoders with weights from the SigLIP2 backbone provides significantly better performance for our following tasks. We then show that the performance of mask segmentation already improves by just using the boundary branch as an auxiliary task, demonstrating the necessity of factoring in the boundary when segmenting liquids. Finally, we show that incorporating cross-attention between the mask branch and the boundary branch further enhances segmentation performance by enabling the mask branch to integrate boundary features effectively. With this, our full network is able to attain the highest performance on all four metrics.

\textbf{Impact of parameter $\omega$}. This parameter is used to balance loss between the mask  and boundary branches. Different values of $\omega$ are tested in Table~\ref{tab:ablation-omega}, and we find that performance levels out at values beyond 200.

\begin{table}[!t]
\caption{Performance for varying values of parameter $\omega$.}
\label{tab:ablation-omega}
\centering
\begin{tabular}{l|cccc}
\toprule
$\omega$ & mIoU & mAcc & mPrec & mF1 \\
\midrule
1.0 & 54.56 & 68.06 & 70.61 & 66.79 \\
10.0 & 54.88 & 67.80 & 72.18 & 66.86 \\
100.0 & 56.01 & 69.50 & 73.98 & 68.17 \\
200.0 & 59.28 & 71.61 & 76.63 & 71.50 \\
\bottomrule
\end{tabular}
\end{table}

\subsection{Segmentation Outside of Liquids}

To test generalization beyond liquids, we evaluate \dm{} on ADE20K~\cite{ADE20K}, a standard semantic segmentation benchmark with 20K images and 150 classes. As shown in Table~\ref{tab:ade20k}, \dm{} achieves a mean IoU of 59.6\%. While ADE20K demonstrates generalization beyond liquids, evaluation on additional robotic perception and segmentation benchmarks remains future work.

\begin{table}[!t]
\caption{Performance of \dm{} in general object segmentation on the ADE20K benchmark.}
\label{tab:ade20k}
\centering
\begin{tabular}{l|c}
\toprule
Method & mIoU \\
\midrule
Mask2Former & 58.9 \\
EoMT        & 58.4 \\
LQDM        & 59.6 \\
\bottomrule
\end{tabular}
\end{table}

\section{Conclusion}

In this work, we introduced LQDS, the first dataset for general liquid segmentation, and LQDM, a dual-branch architecture to enhance mask predictions with boundary features. Experiments show that LQDM outperforms state-of-the-art methods on the testing set of LQDS, establishing a strong baseline for the liquid segmentation task which remains largely unsolved. 
Future work should expand the scale of LQDS, improve representation of rare liquid classes, and further investigate deployment efficiency of models on robotic hardware.
We believe that LQDS and LQDM will facilitate future research in liquid segmentation and enable practical applications in robotics, where the ability to detect and avoid liquids is crucial for safe navigation and manipulation.


\bibliographystyle{IEEEtran}
\bibliography{IEEEabrv, main}

\end{document}